\def\year{2022}\relax
\documentclass[letterpaper]{article} 
\usepackage{aaai22}  
\usepackage{times}  
\usepackage{helvet}  
\usepackage{courier}  
\usepackage[hyphens]{url}  
\usepackage{graphicx} 
\usepackage{natbib}  
\usepackage{caption} 
\DeclareCaptionStyle{ruled}{labelfont=normalfont,labelsep=colon,strut=off} 
\usepackage{algorithm}
\usepackage{algorithmic}

\usepackage{newfloat}
\usepackage{listings}

\usepackage{adjustbox}
\usepackage{multirow}
\usepackage{bbm}
\usepackage{mathtools}
\usepackage{booktabs}
\usepackage{amssymb}
\usepackage{soul}
\usepackage{color}
\usepackage{subfigure}

\floatstyle{ruled}
\newfloat{listing}{tb}{lst}{}
\floatname{listing}{Listing}
\title{Learning from Mistakes - A Framework for Neural Architecture Search}
\author{
    Bhanu Garg\textsuperscript{\rm 1}\equalcontrib, Li Zhang\textsuperscript{\rm 2}\equalcontrib, Pradyumna Sridhara\textsuperscript{\rm 1}, Ramtin Hosseini\textsuperscript{\rm 1}, \\ 
    Eric Xing\textsuperscript{\rm 3}, Pengtao Xie\textsuperscript{\rm 1}
    \\
}
\affiliations{
    \textsuperscript{\rm 1} University of California, San Diego,  USA\\

    \textsuperscript{\rm 2} Zhejiang University, China 
    
    \textsuperscript{\rm 3} Carnegie Mellon University, USA; Mohamed bin Zayed University of Artificial Intelligence, UAE
}

\usepackage{bibentry}

\begin{document}

\maketitle

\begin{abstract}\label{abstract}
Learning from one's mistakes is an effective human learning technique where the learners focus more on the topics where mistakes were made, so as to deepen their understanding. In this paper, we investigate if this human learning strategy can be applied in machine learning. We propose a novel machine learning method called Learning From Mistakes (LFM), wherein the learner improves its ability to learn by focusing more on the mistakes during revision. We formulate LFM as a three-stage optimization problem: 1) learner learns; 2) learner re-learns focusing on the mistakes, and; 3) learner validates its learning. We develop an efficient algorithm to solve the LFM problem. We apply the LFM framework to neural architecture search on CIFAR-10, CIFAR-100, and Imagenet. Experimental results strongly demonstrate the effectiveness of our model.
\end{abstract}

\section{Introduction}\label{intro}
Over the years, humans have accumulated a lot of practical learning techniques. One such effective learning method is to learn from previous mistakes. Initially, the learner learns a concept and evaluates themselves through a test to measure their level of understanding. The topics in the concept where the learner makes more mistakes can be identified as not having been learned well by the learner. Therefore, the learner will re-study the  topic while focusing on the topics where mistakes were made. This will ensure that repetition of similar mistakes in the future is avoided while also strengthening previously well-learned topics. Inspired by this human learning technique, we propose a methodology that can apply the idea of learning from mistakes to machine learning.

With the deep learning era kicking off in machine learning, state-of-the-art neural network performance is achieved mainly by architectures designed manually by experts. The neural architecture search (NAS) and evaluation by human experts require substantial effort and may not give the most optimal performance. Recently, there has been growing interest in automating the manual process of learning architecture design. This paper proposes a general framework that draws inspiration from human learning skills and can be applied to any differentiable architecture search algorithm. We also explore the efficacy of our learning method on NAS.

In our framework, the model consists of two sets of network weights sharing a common learnable architecture, a learnable data encoder, and a coefficient vector - representing different components of the learner. The two sets of network weights correspond to two parts of the main task learning faculties of the learner. The learnable architecture corresponds to skill learning faculty of the learner. The data encoder and coefficient vector correspond to auxiliary faculties of the learner, helping it to recognise and summarize the information respectively. We propose a multi-level optimization framework that uses the above sets of parameters to learn better neural architectures. 

We begin by training the first set of network weights on a training dataset. We then see what mistakes our model makes while predicting on the validation set. Then, for each training example, we assign specific weights based on the mistakes made by the model and the similarity of this example to an incorrectly predicted validation example. 
The second set of network weights are trained on these weighted examples, essentially making the model learn from its mistakes and correct them. The vanilla approaches in NAS do not factor in the variations in learning difficulty. In contrast, our method re-weights the training examples at each stage based on the current capability of the learner, enabling the learner to deal with more challenging cases in the unseen data. Then finally, the architecture, encoder, and coefficient vectors are learned on the second model's validation performance. 


The major contributions of the paper are as follows:
\begin{itemize}
    \item Inspired by the human learning technique, we propose a novel approach to apply the Learning from Mistakes (LFM) method in machine learning.
    \item We formulate LFM as a multi-level optimization framework that involves three stages of learning: learner learns; learner corrects its mistakes; learner, encoder, and coefficient vector validate themselves. 
    \item We apply our approach to neural architecture search on CIFAR-100, CIFAR-10, and ImageNet. The results demonstrate the effectiveness of our method.
\end{itemize}

\section{Related Works}

\subsection{Data Re-weighting and Selection}
Several approaches have been proposed for data selection. Matrix column subset selection~\cite{deshpande2010efficient,boutsidis2009improved} aims to select a subset of data examples that can best reconstruct the entire dataset. Similarly, Coreset selection~\cite{bachem2017practical} chooses representative training examples such that models trained on the selected examples have comparable performance with those trained on all training examples. These methods perform data selection and model training separately. As a result, the validation performance of the model cannot be used to guide data selection. \cite{ren2018learning} proposes a meta-learning method to learn the weights of training examples by performing a meta gradient descent step on the weights of the current mini-batch of examples. \cite{shu2019meta} propose a method that can adaptively learn an explicit weighting function
directly from data. 

Our work takes inspiration from~\cite{ren2018learning} to use a meta gradient step on the weights, and from~\cite{shu2019meta} to explicitly learn a weighting function and extends them to use not only validation and training performance but also the similarity of data and labels. We make this possible by learning the weights and the learner in different stages of the optimization problem.  Moreover, the previous works focus on selecting training (finetuning) examples using a  bi-level optimization framework, while our work focuses on selecting pretraining examples using a three-level optimization framework.

\subsection{Neural Architecture Search (NAS)} 
Recently, NAS has come to the forefront of deep learning techniques due to its success in discovering neural architectures that can substantially outperform manually designed ones. Early versions of NAS such as \cite{zoph2016neural,pham2018efficient,zoph2018learning} used computationally intensive approaches like reinforcement learning -  where the accuracy of the validation set was defined as the reward, and a policy network was trained to generate architectures that can maximize these rewards. Another contemporary approach \cite{liu2018hierarchical,Real_Aggarwal_Huang_Le_2019} was using evolutionary learning techniques - where the set of all architectures represent a population and the fitness score is the validation accuracy of each architecture. Architectures with lower fitness scores would be replaced with higher fitness score architectures. However, even this approach was computationally intensive. To address this problem, differentiable architecture search techniques were explored \cite{cai2018proxylessnas,liu2018darts,xie2018snas} and their results are much more promising because of the use of weight-sharing techniques and the application of gradient descent in a continuous architecture search space. 

DARTS~\cite{liu2018darts} made the first breakthrough in the area of Differentiable NAS. Several other DARTS-based techniques \cite{chen_progressive_2019,xu_pc-darts_2019,liang2019darts+,chu2021darts} have explored to further reduce the cost of computation for differentiable NAS. Some of the approaches include - 
PDARTS~\cite{chen_progressive_2019} increasing the depth of architectures progressively during the searching,  
PC-DARTS~\cite{xu_pc-darts_2019} evaluating only a sub-set of channels and thereby reducing the redundancy in the search space.  

The LFM framework proposed in this paper can be applied to any differentiable NAS method for further enhancement.

\section{Method} \label{sec:method}
In this section, we propose a framework that can imitate human learning in the form of Learning from Mistakes (LFM) and present an optimization algorithm for solving the problem of LFM.

\subsection{ The Learning from Mistakes framework} \label{framework}
%
%
%
%


The framework contains two sets of
network weights $W_1$ and $W_2$ - that are two parts of the same learner and are trying to learn to perform the same target task. They share a learnable architecture $A$. The primary goal here is to help the learner correct the mistakes (made when studying for the first time) during the revision. Further, to help map the topics in the test to topics in the syllabus, there is an encoder with pre-defined neural architecture (by human experts) with learnable network weights $V$; and a coefficient vector $r$. We organized the learning into three stages.

\paragraph{Stage I.} In the first stage, we train the first set of network weights $W_1$ by minimizing the loss on the training dataset $D^{(tr)} $.
The optimal weights $W_{1}^{*}(A)$ is a function of architecture A, which at this stage is fixed, and hence: 
$$W_1^*(A) = \mathop{\arg\min}_{W_1} L(A, W_1, D^{tr})$$
The architecture $A$ is used to define the training loss but is not updated at this stage. If we were to directly learn $A$ by minimizing this training loss, a trivial solution would be yielded where $A$ is very large and complex, and would perfectly overfit the training data but generalize poorly on unseen data.

\paragraph{Stage II.} In the second stage, the goal is to re-weight the training samples for training the second set of network weights $W_2$ of the learner. We apply $W_{1}^{*}(A)$ to the validation dataset $D^{(val)}$ and check its performance on the validation examples. Without loss of generality, we assume the task is image classification. To make the model re-learn whilst focusing more on mistakes on the validation examples by $W_{1}^{*}(A)$, we re-weight each training example $d^{(tr)}_i$ based on the following metrics:

\begin{itemize}
  \item Visual similarity between $d^{(tr)}_i$ and $d^{(val)}_j$, denoted by $x_{ij}$
  \item Label similarity of $d^{(tr)}_i$ and $d^{(val)}_j$, denoted by $z_{ij}$
  \item Valid performance of $W_{1}^{*}(A)$ on $D^{(val)}$, denoted by $u_j$
\end{itemize}

In human learning, a question that has been incorrectly learned can be corrected during the relearning stage by focusing more on examples that are similar to the wrongly learned question. Here, the metric $x_{ij}$ tries to measure how similar a previously incorrectly predicted $d^{(val)}_j$ is to a training example $d^{(tr)}_i$ and $z_{ij}$ depicts whether they describe the same topic. The term $u_j$ measures by how much $W_{1}^{*}(A)$ is wrong for each validation example $j$.

We use these re-weighted training examples to train $W_2$. This allows $W_2$ to focus on the topics that $W_1$, after training, could not get right. 

Visual similarity measures how similar the training example $d^{(tr)}_i$ is to each validation example $d^{(val)}_j$. Let V denote an image encoder. For
each validation example $d^{(val)}_j$, its similarity with  $d^{(tr)}_i$ is defined as the dot-product attention~\cite{luong-pham-manning:2015:EMNLP}  as:
\begin{equation}
\label{visual_similarity}
x_{ij} =  \frac{exp(V(d^{(tr)}_i). V(d^{(val)}_j))}{\sum_{k=1}^{N^{(val)}}exp(V(d^{(tr)}_i) . V(d^{(val)}_k))}  
\end{equation}

where $N^{(val)}$ is the number of validation examples.  $V (d)$
denotes the $K$-dimensional visual representation of the data example d. 

Label similarity measures the similarity between the label of the training example and the label of each validation
example. Let $z_{ij}$ denote the label similarity between  a  validation example $d^{(val)}_j$
and a training example $d^{(tr)}_i$. We
define $z_{ij}$ as:  
\begin{align} \label{z_ij}
    z_{ij} = \mathbb{I}\{y^{(tr)}_i=y^{(val)}_j\}
\end{align} 
where $y$ is the label of the corresponding data example, and $\mathbb{I}\{\}$ the indicator function on the condition being true or not. 

The validation performance $u_j$ of $W_{1}^{*}(A)$ on a validation example $d^{(val)}_j$ is the cross
entropy loss on this example:
\begin{equation} \label{val_perform}
u_j = \text{crossentropy} (f(d^{(val)}_j; W_{1}^{*}(A)), y^{(val)}_j )
\end{equation}
where $f(d^{(val)}_j; W_{1}^{*}(A))$ is the predicted probabilities of $W_{1}^{*}(A)$ on $d^{(val)}_j$ and $y^{(val)}_j$ is the class label of $d^{(val)}_j$. 

Let $x_i$, $z_i$, and $u$ be $N^{(val)}$-dimensional vectors where the $j$-th
element is $x_{ij}$ , $z_{ij}$ , and $u_j$ defined before. We calculate the
weight $a_i$ of the training example $d^{(tr)}_i$ as:
\begin{equation} \label{ai}
    a_i = \text{sigmoid}((x_i \odot  z_i \odot u)^T r)
\end{equation}
where $\odot$ denotes element-wise multiplication, and $r$ is a coefficient vector. Note $a_i$ is a function of $V$, $W_{1}^{*}(A)$, and $r$. 

Given the weight $a_i$ of each training example, we train the second set of network weights $W_2$ by minimizing the weighted
training loss, with the architecture $A$, encoder $V$, and $r$ fixed. 
\begin{align} \label{w2_eqn}
    W_2^*(W_1^*(A), V, r) = \mathop{\arg\min}_{W_2} \sum_{i=1}^{N^{tr}} a_i \ell(A, W_2(A), d_i^{(tr)})
\end{align}

\paragraph{Stage III.} In the third and final stage,
the encoder $V$, coefficient vector $r$, and the architecture  $A$ minimise the validation loss of $W_2^*(W_1^*(A), V, r)$.
\begin{align} \label{avr_eqn}
    A, V, r = \mathop{\arg\min}_{A,V,r} L(A, W_2^*(W_1^*(A), V, r), D^{(val)})
\end{align}

Putting the above pieces together, we have the following LFM framework, which is a three level optimization problem:

\begin{align}\label{optimisation_eqn}
 & A, V, r  =  \mathop{\arg\min}_{A,V,r} L(A, W_2^*(W_1^*(A), V, r), D^{(val)})  \nonumber \\
  & s.t. \quad W_2^*(A) = \mathop{\arg\min}_{W_2} \sum_{i=1}^{N^{tr}} a_i \ell (A, W_2(A), d_i^{(tr)})  \\
  & \qquad \ W_1^*(A) = \mathop{\arg\min}_{W_1} L(A, W_1, D^{tr}) \nonumber
\end{align}

\begin{figure}[ht]
    \centering
    \includegraphics[width=7cm, height=6cm]{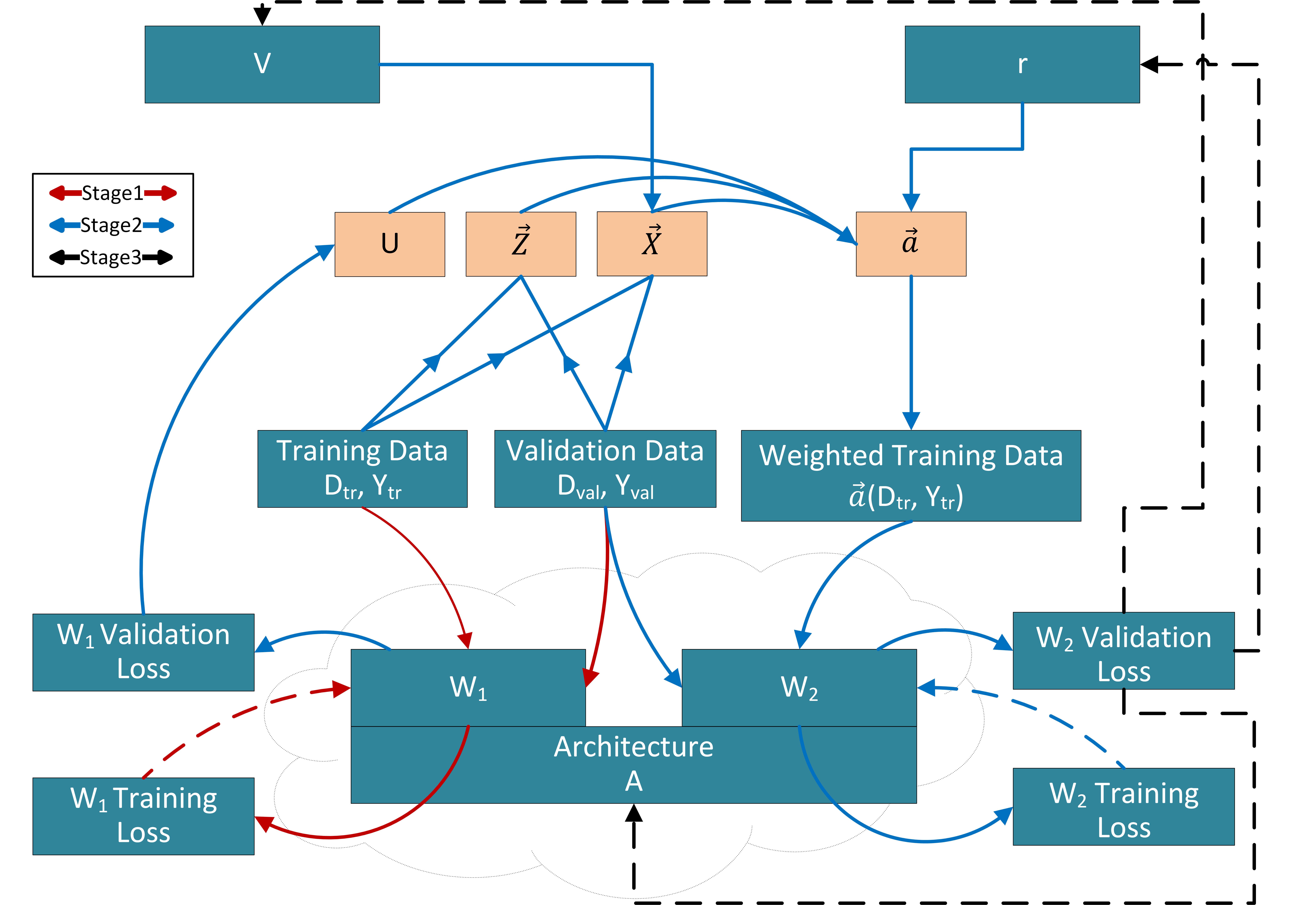}
    \caption{The overall process flow of our method when applying to NAS. The red arrows indicate stage 1 processes, blue arrows indicate stage 2 processes, and black arrows indicate stage 3 processes. 
    }
    \label{fig:2}
\end{figure}

We summarise the above equations in the process flow diagram Figure \ref{fig:2}. In our end-to-end framework, $W_1$ will learn from its previous mistakes and avoid making similar mistakes, in an indirect way. After the weights of $W_2$ are trained by correcting the mistakes made by $W_1$, the architecture  will be updated accordingly since the gradient of $A$ depends on 
$W_2$; an updated $A$  will render $W_1$
 to change as well since the gradient of $W_1$ 
 depends on $A$ . Along the chain $W_2 \xrightarrow[]{} A \xrightarrow[]{} W_1$ 
 chain, $W_1$ is indirectly influenced by 
 $W_2$. 

Similar to~\cite{liu2018darts}, we represent the architecture $A$ of the learner in a differentiable way. The search space of $A$ is composed of many building blocks, where the output of each block is associated with a weight $a$ indicating the importance of the block. After learning, block whose weight $a$ is among the largest are retained to form the final architecture. To this end, architecture search amounts to optimizing the set of architecture weights $A = \{a\}$. Our framework can be applied on top of any differentiable architecture search methods such as DARTS~\cite{liu2018darts}, PDARTS~\cite{chen_progressive_2019}, PC-DARTS~\cite{xu_pc-darts_2019}, DARTS-~\cite{chu2021darts} among others.

\subsection{Optimization Algorithm}\label{optimisation_algo}

In this section, we derive an optimization algorithm to solve the LFM problem defined in Equation \ref{optimisation_eqn}. Inspired by \cite{liu2018darts}, we approximate $W_1^*(A)$ and $W_2^*(W_1^*(A), A, V, r)$ one step gradient descent updates for the inner optimization equations to reduce the computational complexity.  

\paragraph{Stage I.} For Stage 1, we approximate $W_1^{*}(A)$ using one step descent for the loss on training data $L(A, W_1, D^{tr})$ keeping $A$ constant as follows:
\begin{align}\label{st1_approx}
    W_1'(A) = W_1 - \eta_{W_1} \nabla_{W_1} L(A, W_1, D^{tr})
\end{align}

\paragraph{Stage II.} For Stage 2, we use $W_1'$ from the Equation \ref{st1_approx} to get  $u_j$. We compute $x_i$ and $z_i$ for each training sample $d_i^{tr}$ using the equations \ref{visual_similarity} and \ref{z_ij}, to finally compute $a_i$ as:
\begin{align} \label{ai_cal:final}
    a_i(d^{val}, d_i^{tr}) &=  \text{sigmoid}((x_i \odot  z_i \odot u)^T r)
\end{align}

Next, we use one step gradient descent to approximate $W_2^*(A,W_1^*(A),V,r)$ as:
\begin{align} \label{st2_approx}
    W_2'(A) = W_2 - \eta_{W_2}  \nabla_{W_2} \sum_{i=1}^{N^{tr}} a_i  \ell (A, W_2, d_i^{(tr)})
\end{align}

\paragraph{Stage III.} For Stage 3, we plug Equation \ref{st2_approx} to learn  architecture $A$ , encoder $V$, and coefficient vector $r$ from the validation loss $L(A, W_2'(W_1'(A), V, r), D^{(val)})$. 
The encoder model $V$ can be updated as: 
\begin{align} \nonumber
 V' &= V - \eta_V \nabla_V L(A, W_2', D^{(val)}) \\ \nonumber
& = V + \eta_V \eta_{W_2} \frac{\partial}{\partial V} \Big(\big( \nabla_{W_2}\sum_{i=1}^{N^{tr}} a_i(V) \ell(A, W_2, d_i^{(tr)})\big) . \nonumber \\
 & \qquad \qquad \qquad \qquad \big( \nabla_{W_2'} L(A, W_2',D^{(val)})\big)\Big) \label{v_updt}
\end{align}

Similarly, the coefficient vector $r$ can be updated as: 
\begin{align} \nonumber
 r' &= r - \eta_r \nabla_r L(A, W_2', D^{(val)}) \\\nonumber
 &= r  + \eta_r \eta_{W_2} \frac{\partial}{\partial r} \Big(\big( \nabla_{W_2}\sum_{i=1}^{N^{tr}} a_i(V) l(A, W_2, d_i^{(tr)})\big). \nonumber \\ 
 &\qquad \qquad \qquad \qquad \big( \nabla_{W_2'} L(A, W_2',D^{(val)})\big)\Big)
 \label{r_updt}
\end{align}

Next, we give the update equation for the  architecture $A$.
\begin{align} \nonumber
 A' &= A - \eta_A \nabla_A L(A, W_2'(A), D^{(val)}) \\ \nonumber
  &= A  - \eta_A \big(\frac{\partial L(A, W_2',D^{(val)})}{\partial A} + \nonumber \\ 
 & \qquad \qquad \qquad \frac{\partial L(A, W_2',D^{(val)})} {\partial W_2'} \frac{\partial W_2'(A)}{\partial A}\big)  \label{updateA}
\end{align}

To save space, the complete derivation is not given in the paper. 
\begin{align} \nonumber
    &A' =  A  - \eta_A \big( \frac{\partial L(A, W_2',D^{(val)})}{\partial A} -  \eta_{W_2} \big(\\ \nonumber
    & -\eta_{W1} \frac{\nabla_{A}L(A, W_1^+, D^{(tr)}) - \nabla_{A}L(A, W_1^-, D^{(tr)})}{2 \epsilon_1} \nonumber \\ 
    & +  \frac{\nabla_A \sum_i a_i \big( L(A, W_2^+, d_i^{(tr)}) - L(A, W_2^-, d_i^{(tr)})\big) }{2\epsilon_2} \big) \big) \label{A_updt}
\end{align}

where  \begin{align*}W_1^\pm &= W_1 \pm \epsilon_1 \nabla_{W_1'}  \big((\nabla_{W_2} \sum_{i=1}^{N^{(tr)}} \textbf{a}_i(W_1') l(A, W_2, d_i^{(tr)})). \nonumber \\
    & \qquad \nabla_{W_2'}L(A, W_2',D^{(val)})\big) \nonumber
\end{align*} and 
\begin{align*}
W_2^\pm = W_2 \pm \epsilon_2 \nabla_{W_2'} L(A, W_2',D^{(val)}) \end{align*} 
Also,  $\epsilon_1$ and   $\epsilon_2$ are small scalars.

The overall algorithm of LFM when applying to other related methods can be summarised in Algorithm \ref{algo1}. And LFM can be applied to any differentiable NAS methods. To apply our framework to a differentiable NAS method $M$, we just need to set the architecture variable $A$ in our framework to be the search space of $M$. In other words, $M$ is a special solution in the solution space $S$ of our method.

\begin{algorithm}[H]
\begin{algorithmic}[1]
\WHILE{not converged}
    \STATE Update $W_1$
    \STATE Update $W_2$
    \STATE Update $A, V, r$ respectively
\ENDWHILE
\caption{Algorithm for LFM}
\label{algo1}
\end{algorithmic}
\end{algorithm}

\section{Experiments} \label{sec:experiments}

\subsection{Datasets}
The experiments are performed on three popular NAS datasets, namely CIFAR-10, CIFAR-100~\cite{krizhevsky_learning_nodate} and ImageNet~\cite{deng2009imagenet}. CIFAR-10 contains $10$ classes, and CIFAR-100 contains $100$ classes. Both these datasets contain 60K images each with each class having the same number of images. We split each of these datasets into a training set with 25K images, a validation set with 25K images, and a test set with 10K images. During architecture search in LFM, the training set is used as $D^{tr}$ and the validation set is used as $D^{val}$. During architecture evaluation, the learned network is trained on the combination of $D^{tr}$ and $D^{val}$. Further, ImageNet contains 1.2M training images and 50K test images with 1000 objective classes.  

\subsection{Experimental Settings}
Our framework is orthogonal to existing NAS approaches and can be applied to any differentiable NAS method. In our experiments, we applied LFM to DARTS~\cite{liu2018darts} and PDARTS~\cite{chen_progressive_2019}. The search spaces of these methods are composed of (dilated) separable convolutions with sizes of 3 × 3 and 5 × 5, max pooling with the size of 3 × 3, average pooling with the size of 3 × 3, identity, and zero. For the encoder $V$, Res-Nets pre-trained on Imagenet were used. Each LFM experiment was repeated three times with different random seeds. The mean and standard deviation of classification errors obtained from the three runs are reported.

\begin{table}[t]
\small
    \centering
    \begin{tabular}{l|ccc}
    \toprule
    Method & Error(\%)\\
    \midrule
    *ResNet \cite{he2016deep}&22.10\\
     *DenseNet \cite{HuangLMW17}&17.18\\
    \hline
    *PNAS \cite{LiuZNSHLFYHM18}&19.53\\
    *ENAS \cite{pham2018efficient}&19.43\\
    *AmoebaNet \cite{real2019regularized}&18.93\\
    \hline
    *GDAS \cite{DongY19}&18.38\\
    *R-DARTS \cite{ZelaESMBH20}&18.01$\pm$0.26
    \\
       *DARTS$^{-}$ \cite{abs-2009-01027}&17.51$\pm$0.25\\
       *DARTS$^{-}$ \cite{abs-2009-01027}& 18.97$\pm$0.16\\
       *DARTS$^{+}$ \cite{abs-1909-06035}&17.11$\pm$0.43\\
      *DropNAS \cite{HongL0TWL020} & 16.39\\
      $\;\;$Random search & 21.92$\pm$0.34 \\
      $\;\;$Random sampling & 21.37$\pm$0.48 \\
\hline
     \hline
            *DARTS-2nd \cite{liu2018darts}  & 20.58$\pm$0.44\\
             $\;\;$LFM-DARTS-2nd-R18 (ours)  & \textbf{17.65$\pm$0.45}\\
              \hline
            *PDARTS \cite{chen_progressive_2019}&17.49\\
       $\;\;$LFM-PDARTS-R18 (ours)  & \textbf{16.44$\pm$0.11}\\
       $\;\;$LFM-PDARTS-R34 (ours)  &  \textbf{15.69$\pm$0.15}\\
        \bottomrule
    \end{tabular}
    \caption{Test error on CIFAR-100. LFM-DARTS-2nd-R18 represents that LFM is applied onto DARTS-2nd, and ResNet-18 is used as the data encoder. Similar meanings hold for other notations in such a format. Results marked with * are obtained from Skillearn \cite{xie_skillearn_2020}. The information about parameters and search cost are shown in the supplement.}
    \label{tab:cifar100}
\end{table}

\paragraph{Architecture Search} During architecture search for CIFAR-10 and CIFAR-100, the architectures of $W_1$ and $W_2$ are a stack of 8 cells. Each cell consists of 7 nodes. We set the initial channel number to 16. For the architecture of the encoder model, we experimented with ResNet-$18$ and ResNet-$34$~\cite{7780459}. The search algorithm was based on SGD, and the hyperparameters of epochs, initial learning rate, and momentum follow the original implementation of the respective DARTS~\cite{liu2018darts} and PDARTS~\cite{chen_progressive_2019}.  We use a batch size of 64 for both DARTS and PDARTS. LFM-PDARTS uses first order approximations to be consistent with the original implementation in PDARTS~\cite{chen_progressive_2019}. The LFM experiments in this paper use A100 for DARTS and A40 for PDARTS. 

\paragraph{Architecture Evaluation} During architecture evaluation for CIFAR-10 and CIFAR-100, a larger network of each category-specific model is formed by stacking 20 copies of the searched cell. The initial channel number is set to 36. We trained the network with a batch size of 96, an epoch number of 600, on a single  Tesla v100 GPU.  On ImageNet, we evaluate the architectures searched on CIFAR-10 or CIFAR-100. In either type, 14 copies of optimally searched cells are stacked into a large network, which was trained using two Tesla A100 GPUs on the 1.2M training images, with a batch size of 1024 and an epoch number of 250. The initial channel number is set to 48.

The LFM method is used to learn the architecture $A$, while the weights $W_1$, $W_2$, $V$, and $r$ learnt during the LFM search are discarded during the architecture evaluation. All the architecture evaluations are run using the same standardized setup as described in the above paragraph. This results in a fair comparison between architectures learnt from different methods, as all the models during evaluation have same number of parameters and hyper-parameters such as epochs, learning rate, and batch size.

\subsection{Results}

\begin{table}[t]
\small
    \centering
    \begin{tabular}{l|ccc}
    \toprule
    Method& Error(\%)\\
    \midrule
    *DenseNet
    \cite{HuangLMW17}&3.46\\
    \hline
     *HierEvol \cite{liu2017hierarchical}&3.75$\pm$0.12\\
        *PNAS \cite{LiuZNSHLFYHM18} &3.41$\pm$0.09\\
    *NASNet-A \cite{zoph2018learning} & 2.65\\
    *AmoebaNet-B \cite{real2019regularized} & 2.55$\pm$0.05\\
    \hline
        *R-DARTS \cite{ZelaESMBH20} &2.95$\pm$0.21\\ 
        *GTN~\cite{abs-1912-07768}& 2.92$\pm$0.06\\
        *BayesNAS \cite{ZhouYWP19} &2.81$\pm$0.04\\
        *MergeNAS \cite{WangXYYHS20} &2.73$\pm$0.02\\
        *NoisyDARTS \cite{abs-2005-03566} &2.70$\pm$0.23\\
            *ASAP \cite{NoyNRZDFGZ20} &2.68$\pm$0.11\\
                *SDARTS
    \cite{abs-2002-05283}&2.61$\pm$0.02\\
            *DropNAS \cite{HongL0TWL020} &2.58$\pm$0.14\\
       *DrNAS \cite{abs-2006-10355} &2.54$\pm$0.03\\
       $\;\;$Random search & 3.07$\pm$0.17 \\
      $\;\;$Random sampling & 2.75$\pm$0.09 \\
    \hline
        \hline
               *DARTS-2nd \cite{liu2018darts} &2.76$\pm$0.09\\
           $\;\;$LFM-DARTS-2nd-R18 (ours)   &\textbf{2.70$\pm$0.06}\\
                       \hline
    *PDARTS \cite{chen_progressive_2019}& 2.50\\
     $\;\;$LFM-PDARTS-R18 (ours) &\textbf{2.46$\pm$0.04}\\
        \bottomrule
    \end{tabular}
    \caption{
    Test error on CIFAR-10. Results marked with * are obtained from Skillearn~\cite{xie_skillearn_2020}. The other notations are same as described in Table \ref{tab:cifar100}. The information about parameters, search cost and more compared methods are shown in the supplement. 
    }
    \label{tab:cifar10}
\end{table}

\begin{table*}[t]
    \centering
    \begin{tabular}{l|cccc}
    \toprule
  \multirow{ 2}{*}{Method}   & Top-1  &Top-5 \\
         & Error (\%) & Error (\%)\\
    \midrule
    *Inception-v1 \cite{googlenet}&30.2 &10.1\\
    *MobileNet \cite{HowardZCKWWAA17} &  29.4& 10.5\\
    *ShuffleNet 2$\times$ (v1) \cite{ZhangZLS18} &  26.4 &10.2 \\
    *ShuffleNet 2$\times$ (v2) \cite{MaZZS18} &  25.1 &7.6\\
    \hline
    *NASNet-A \cite{zoph2018learning} &26.0 &8.4\\
    *PNAS \cite{LiuZNSHLFYHM18} &25.8 &8.1\\
    *MnasNet-92 \cite{TanCPVSHL19} & 25.2 & 8.0\\
        *AmoebaNet-C \cite{real2019regularized} &  24.3 &7.6\\
    \hline
     *SNAS-CIFAR10 \cite{xie2018snas} & 27.3 &9.2\\
                    *PARSEC-CIFAR10 \cite{abs-1902-05116} & 26.0 &8.4\\
                 *DSNAS-ImageNet \cite{HuXZLSLL20} &25.7& 8.1\\
          *SDARTS-ADV-CIFAR10 \cite{abs-2002-05283}&25.2& 7.8\\
     *FairDARTS-ImageNet \cite{abs-1911-12126} &24.4 &7.4\\
             *DrNAS-ImageNet \cite{abs-2006-10355} & 24.2 &7.3\\
             *ProxylessNAS-ImageNet \cite{cai2018proxylessnas} & 24.9 &7.5  \\
             *GDAS-CIFAR10 \cite{DongY19} &  26.0&8.5\\
     \hline
       \hline
       *DARTS2nd-CIFAR10 \cite{liu2018darts}  & 26.7 &8.7\\
        $\;\;$LFM-DARTS-2nd-CIFAR10 (ours) &  \textbf{25.12} & \textbf{7.65} \\
        \hline
          *PDARTS (CIFAR10) \cite{chen_progressive_2019}&24.4 &7.4\\
        $\;\;$LFM-PDARTS-CIFAR10 (ours) & \textbf{24.14}&\textbf{6.85}\\
        \hline
             *PDARTS (CIFAR100) \cite{chen_progressive_2019}&24.7& 7.5\\
           $\;\;$LFM-PDARTS-CIFAR100 (ours)& \textbf{24.11} & \textbf{6.70}\\
        \bottomrule
    \end{tabular}
    \caption{Top-1 and top-5 classification errors on the test set of ImageNet. Results marked with * are obtained from Skillearn \cite{xie_skillearn_2020}. From top to bottom, on the first three blocks are 1) networks manually designed by humans; 2) non-gradient based NAS methods; and 3) gradient-based NAS methods. Rest of the notations follow Tables \ref{tab:cifar100}, \ref{tab:cifar10}. The information about parameters, search cost and more compared methods are shown in the supplement. 
    }
    \label{tab:imagenet}
\end{table*}

\subsubsection{Result of classification in different datasets}

The results of the classification error(\%) of different NAS methods on CIFAR-100 are showed in Table \ref{tab:cifar100}. We make the following observations from this table: 
\begin{itemize}
    \item When LFM is applied  DARTS-2nd (second-order approximation) and PDARTS, the classification errors of these methods are reduced significantly. For example, when LFM is applied to DARTS-2nd, the error reduces from $20.58\%$ to $17.70\%$. In PDARTS, the error reduces from $17.49\%$ to $16.44\%$ (when using ResNet-18 as encoder $V$). This shows the effectiveness of our method in improving the performance of architecture search. In the baseline NAS approaches, all the training examples have the same weight, which implicitly implies that all examples are equally difficult to learn. The learner, in this case, can give a good performance by performing well on the majority of easy examples and ignoring the minority of difficult examples. In contrast, our method re-weights the training example at each stage based on the current capability of the learner, giving more weight to examples that are difficult to learn, which is a more realistic scenario. The learner gains the ability to deal with more challenging cases in the unseen data. 
    \item LFM-PDARTS-R34 outperforms LFM-PDARTS-R18 by $0.75\%$, where the former uses ResNet-34 as the image encoder, while the latter uses ResNet-18. ResNet-34 is a deeper and more powerful data encoder than ResNet-18. This shows that mapping the validation examples to similar training examples is a core component contributing to the effectiveness of our proposed LFM method.
    \item LFM-PDARTS-R34 achieves the best performance among all methods, which demonstrates the effectiveness of applying LFM to differentiable NAS methods and improving their performance. To the best of our knowledge,  LFM-PDARTS-R34 is the new SOTA on CIFAR-100.
\end{itemize} 

The results of the classification error(\%) of different NAS methods on CIFAR-10 are showed in Table \ref{tab:cifar10}. As can be seen, LFM applied to DARTS-2nd and PDARTS reduces the errors of these baselines by roughly $0.05\%$. This further demonstrates the effectiveness of our method.

The results of the classification error(\%) - top-1 and top-5 of different NAS methods on ImageNet are showed in Table \ref{tab:imagenet}. In methods LFM-DARTS-2nd-CIFAR10 and  LFM-PDARTS-CIFAR10, the architecture searched on CIFAR-10 is evaluated on ImageNet, whereas in LFM-PDARTS-CIFAR100, the architecture searched on CIFAR-100 is evaluated on Imagenet. The LFM-DARTS-2nd-CIFAR10 outperforms the baseline DARTS-2nd-CIFAR10 by $1.6\%$, while LFM-PDARTS-CIFAR100 outperforms its corresponding baseline  by $0.6\%$, and LFM-PDARTS-CIFAR10 by $0.3\%$.  As shown, the LFM methods outperform their corresponding baselines, and therefore, demonstrate our method's effectiveness.

\subsubsection{Ablation Studies}

\begin{figure}[ht]
    \centering
    \includegraphics[height=5cm]{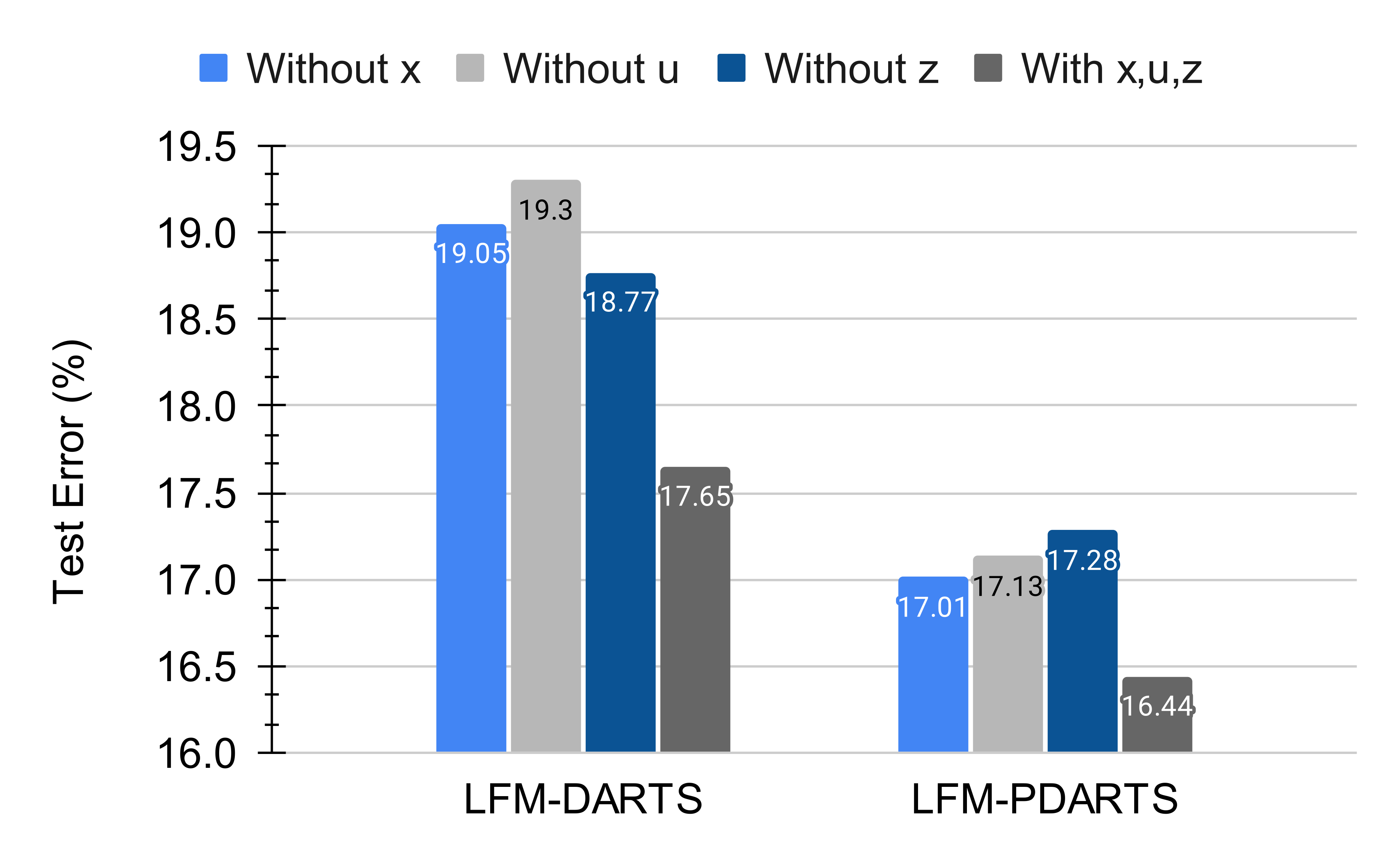}
    \caption{Ablation on components of $a_i$. The left bar column shows the comparison of ablated models 1) without x, 2) without u, 3) without z , 4) and the full models. }
    \label{fig:3}
\end{figure}

\paragraph{Ablation 1} Our method introduces three important components to the re-weighting parameter $a_i$ namely: $x$, $u$, and $z$. In this study, we demonstrate the effect of ablating each component. We perform the experiment of CIFAR-100 and set the encoder to ResNet-18. The other details are are kept same as the base experiments described in earlier sections. The performances of the ablated models are shown in \ref{fig:3}. These results demonstrate the effectiveness and necessity of measuring mistakes $u$ on validation examples, calculating visual similarity $x$ and label similarity $z$ between training and validation examples.

\begin{figure}[ht]
    \centering
    \includegraphics[height=5cm]{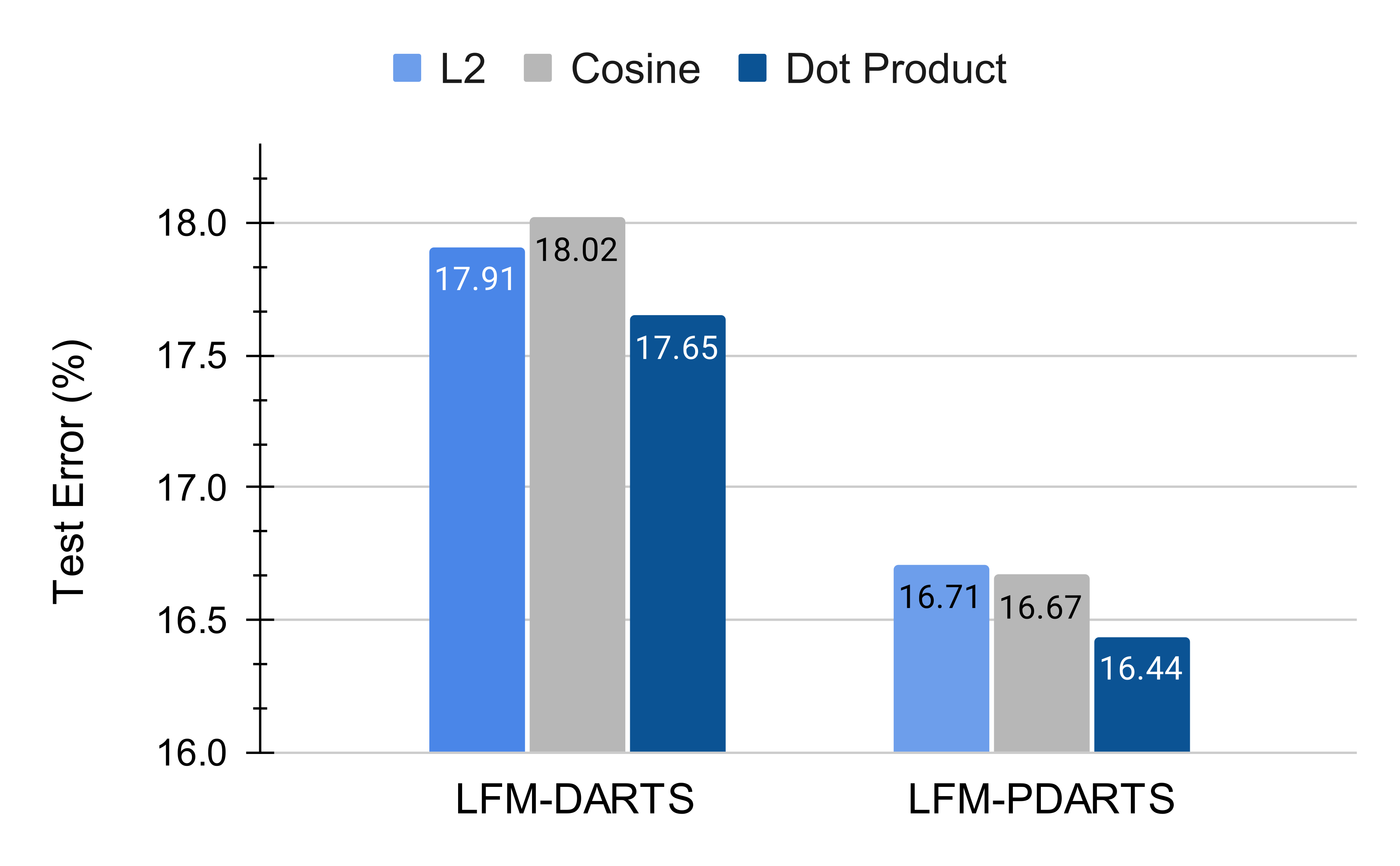}
    \caption{Comparison of different visual similarity metrics.}  
    \label{fig:4}
\end{figure}

\paragraph{Ablation 2} For visual similarity, we use dot product attention, which has shown broad effectiveness in many applications and is simple to use. To demonstrate its effectiveness in LFM, we compared with other metrics such as cosine similarity and L2 distance. The other details are are kept same as the base experiments described in earlier sections. Results are shown in Figure \ref{fig:4}. Dot product attention used in our framework works better than the other two metrics.

\paragraph{Ablation 3} Since we use the data encoder that is pre-trained on ImageNet, it may provide an unfair advantage to our method in the CIFAR100 experiment since the encoder is exposed to more data. We set experiments to restrict our method to only using the CIFAR100 dataset and check its performance. We train the encoder solely on CIFAR-100, without using ImageNet pre-training. The results (on Cifar100) are given in Table \ref{tab:pre} below. Removing ImageNet pretraining does not increase test errors significantly, showing that the improvement achieved by our method over DARTS/PDARTS  is not due to ImageNet pretraining.

\begin{table}[ht]
\begin{center}
\begin{tabular}{l|c}
\hline
Method   &	Test error(\%)\\
\hline
LFM+DARTS, no INP &	17.82$\pm$0.39 \\
LFM+DARTS+INP& 17.65$\pm$0.45 \\
DARTS&	20.58$\pm$0.44\\
\hline
LFM+PDARTS, no INP &		16.51$\pm$0.13 \\
LFM+PDARTS+INP& 16.44$\pm$0.11 \\
PDARTS &17.49\\
\hline
\end{tabular}
\end{center}
\caption{Comparisons on different pre-training datasets of the encoder. InP means ImageNet pretrain.}
\label{tab:pre}
\end{table}

\subsection{Limitations and Future Work}\label{sec:limitations} 

Our method requires the use of two learners that have similar learning capabilities, so that one can learn from the mistakes of other. This increases the memory requirements and makes the learning slow compared to the vanilla approaches. As future work, we explore to reduce memory cost during architecture search by parameter-sharing between the three models $W_1$, $W_2$, and $V$. For $W_1$ and $W_2$, we let them share the same convolutional layers, but have different classification heads. For $V$, we replace ResNet-18 with $W_1$. As shown in Table \ref{tab:memory} that via parameter sharing (PS), the memory and computation costs of our method are reduced to a level similar to vanilla DARTS and PDARTS, while our method still achieves significantly lower test errors than DARTS and PDARTS. A future work direction is to improve memory usage while keeping the full performance of the LFM method. Another direction is to extend the applicability of LFM to other meta-learning tasks such as data re-weighting, or more complex tasks like semantic segmentation. Further, LFM can be extended to language modeling tasks as well. 

\begin{table}[ht]
\centering
\begin{footnotesize}
\begin{tabular}{l|ccc}
\toprule
\multirow{2}{*}{Method} & Test error & Memory & Cost \\
                        & (\%) & (MiB) & (days) \\
\hline
LFM+DARTS, no PS & 17.65$\pm$0.45 & 23,702 & 5.4 \\
LFM+DARTS+PS & 18.77$\pm$0.31 & 12,138 & 1.6 \\
DARTS & 20.58$\pm$0.44 & 11,053 & 1.5 \\
\hline
LFM+PDARTS, no PS & 16.44$\pm$0.11 & 20,744 & 2.0 \\
LFM+PDARTS+PS & 16.83$\pm$0.08 & 10,582 & 0.3 \\
PDARTS & 17.49 & 9,659 & 0.3 \\
\bottomrule 
\end{tabular}
\end{footnotesize}
\caption{Test error(\%), memory cost (MiB) and computation cost (GPU days) of different models on CIFAR-100.}
\label{tab:memory}
\end{table} 

%
\section{Conclusions}

In this paper, we propose a novel strategy to improve the performance of differential Neural Architecture Search approaches - Learning from Mistakes (LFM) inspired by the astounding ability of humans to learn from feedback on their current performance. The learner model continuously focuses more on the mistakes and improves its learning ability on more challenging examples. We propose a multi-level optimization framework to formalize LFM and provide an efficient solution to the problem. We apply LFM to NAS on datasets of CIFAR-10, CIFAR-100, and Imagenet - pushing the frontiers on NAS research.

\bibliography{bib}
\end{document}